\documentclass[11pt,a4paper]{article}
\usepackage[margin=1in]{geometry}
\usepackage[T1]{fontenc}
\usepackage{lmodern}
\usepackage{microtype}
\usepackage{amsmath,amssymb,bm}
\usepackage[authoryear,round]{natbib}
\usepackage{url}
\usepackage{booktabs}
\usepackage{multirow}
\usepackage{graphicx}
\usepackage[hidelinks]{hyperref}
\setcitestyle{citesep={;},aysep={,},yysep={;}}
\title{Where Scientific Search Agents Fail: Decision-Checkpoint Auditing of Exposure and Inspection Attempts}

\author{%
Hongmin Li\textsuperscript{1,2}\qquad Wanli Zhao\textsuperscript{3}\\[0.8em]
\small \textsuperscript{1}School of Life Science and Technology, Institute of Science Tokyo\\
\small Tokyo, Japan\\[0.4em]
\small \textsuperscript{2}Department of Computational Biology and Medical Sciences\\
\small Graduate School of Frontier Sciences, The University of Tokyo\\
\small Kashiwa, Japan\\[0.4em]
\small \textsuperscript{3}Department of Information Science\\
\small Graduate School of Advanced Science and Engineering, Hiroshima University\\
\small Higashihiroshima, Japan\\[0.6em]
\small \texttt{lihongmin@edu.k.u-tokyo.ac.jp}}
\date{}

\begin{document}
\maketitle

\begin{abstract}
Final-answer accuracy provides limited insight into where scientific-search agents fail: a target paper may remain unseen, appear in search results without being inspected, or be inspected before an incorrect answer. We introduce decision-checkpoint auditing, which records search observations and executed tool actions during inference and classifies outcomes by matching these events to target identities and evaluator labels afterward. Across five search conditions on 540 answerable AutoResearchBench Deep questions in a fixed, target-enriched environment, keyword search achieves 24.6\% accuracy, compared with 17.8\% for raw search. Keyword search has fewer failures in which the target is neither exposed nor inspected, but more failures after exposure without inspection. A read-first policy makes 27.4\% more evidence-search calls than keyword search, while target inspection attempts occur on 199 rather than 191 questions. Both policies achieve 24.6\% accuracy, with 60 gains and 60 losses in paired answer outcomes. These results show how trajectory-level measurements distinguish target exposure, inspection attempts and answer correctness, revealing differences obscured by aggregate accuracy and tool-use totals.
\end{abstract}

\section{Introduction}

\begin{figure}[t]\centering
\includegraphics[width=\linewidth]{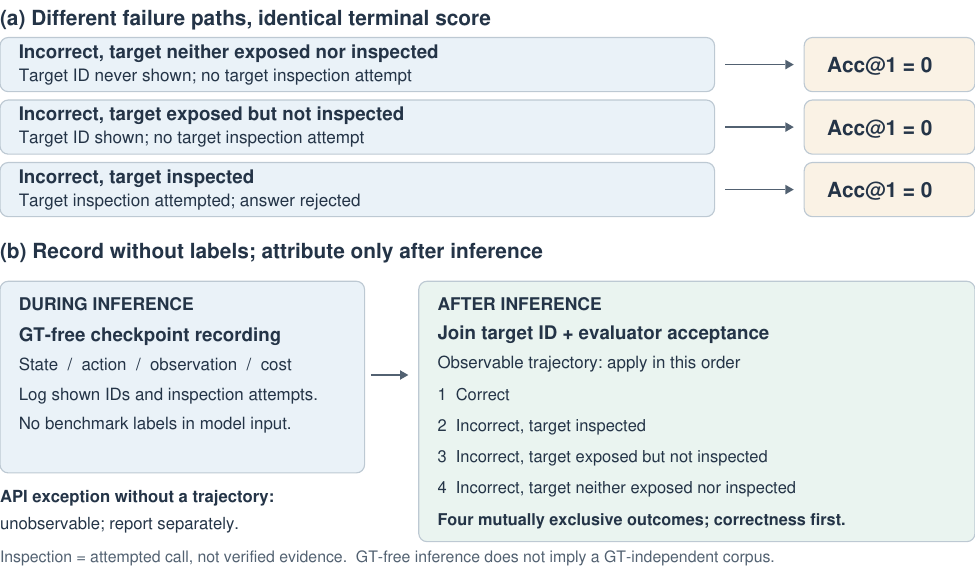}
\caption{Schematic of decision-checkpoint auditing. (a) Three distinct failure paths share the same terminal score. (b) Inference-time checkpoints contain no ground-truth (GT) labels; target identities and evaluator labels are added afterward. Observable trajectories are classified by correctness, then inspection and exposure. API exceptions without trajectories form a separate unobservable category. Inspection denotes an executed within-paper evidence-search attempt.}\label{fig:concept}
\end{figure}

Scientific-search agents identify papers by alternating between search queries, candidate inspection and answer selection~\citep{he2025pasa,autoresearchbench}. A final paper-identification score records whether the selected paper matches the benchmark target, but leaves the intervening search decisions unresolved. An agent may answer incorrectly after failing to expose the target, leaving an exposed target uninspected, or attempting to inspect the target. Distinguishing these events specifies which part of a search trajectory an intervention is intended to change (Figure~\ref{fig:concept}).

Existing work examines intermediate retrieval stages and search failures. PaperQA2 tracks source retention through its evidence pipeline~\citep{skarlinski2024paperqa}, AutoResearchBench manually analyzes scientific-search errors~\citep{autoresearchbench}, and SearchAuditor diagnoses selected failed trajectories~\citep{liang2026searchauditor}. We study a complementary measurement question: across the same benchmark questions, how do target exposure, target inspection attempts and accepted answers differ between search conditions?

Our approach represents search trajectories as \emph{decision checkpoints} and identifies target exposure and inspection attempts by matching search observations and executed actions to benchmark paper IDs after inference. These event labels require no additional semantic judge. Inspection denotes an executed within-paper evidence-search call, whose returned excerpts may or may not provide sufficient evidence for the question. Answer correctness is assessed separately by an adapted benchmark evaluator. Ground-truth labels are withheld during inference, although target identities inform corpus construction.

\begin{samepage}
The paper makes two contributions. First, we define an event-based framework linking target exposure and executed inspection attempts to answer correctness on a common set of questions. Second, we characterize these outcomes across five search conditions on 540 answerable AutoResearchBench Deep questions in a shared, target-enriched environment. Keyword search has fewer failures without target exposure or inspection than raw search, but more failures after exposure without inspection. For read-first, separating agent-initiated and gate-forced calls reveals how inspection effort is distributed across candidate papers. Paired comparisons further reveal changes in target inspection and answer correctness that identical aggregate accuracies conceal.
\par\end{samepage}

\section{Related Work}
\paragraph{Scientific search and stage measurement.} AutoResearchBench defines Deep and Wide literature search over arXiv~\citep{autoresearchbench}. Its Appendix F manually analyzes sampled incorrect predictions, including evidence aggregation, tool execution and candidate ranking failures; the latter explicitly includes a gold paper already in the candidate set but ranked below another paper. PaSa models search/read/select decisions~\citep{he2025pasa}, LitSearch benchmarks retrieval and reranking~\citep{ajith2024litsearch}, and OpenScholar generates citation-supported answers~\citep{asai2024openscholar}. PaperQA2 tracks source-DOI recall through retrieval, chunk ranking, contextual summarization and attribution~\citep{skarlinski2024paperqa}, a direct precedent for separating discovery from downstream evidence use. We measure target exposure and executed inspection attempts across the evaluated questions, linking these events to paired answer outcomes.

\paragraph{Process and evidence evaluation.} AgentBoard measures partial progress through goal-state similarity or defined subgoals~\citep{ma2024agentboard}; ToolSandbox matches temporally ordered milestones involving tool calls and environment states~\citep{lu2025toolsandbox}; AgentEval evaluates DAG-structured steps and error propagation~\citep{agenteval2026}. SearchAuditor evaluates failure localization, attribution and repair against expert annotations of selected failed trajectories, and relates exploration to downstream failure types~\citep{liang2026searchauditor}. Semantic evidence diagnostics include SciFact's rationale and support/refutation labels~\citep{wadden2020scifact} and RAGChecker's claim-level entailment checks for retrieval and generation~\citep{ru2024ragchecker}. Our target-identity and executed-action labels provide an event-level complement to these semantic diagnoses, covering successes as well as failures.

\paragraph{Retrieval access and resource allocation.} BrowseComp-Plus fixes the corpus and varies whole-document reader access and oracle retrieval with annotated positive documents~\citep{chen2025browsecompplus}. Cost-aware evaluation reports accuracy jointly with resource use~\citep{kapoor2024agentsmatter}, while budget-aware control~\citep{budgetcontrol} and value-guided search~\citep{valuetree} optimize allocation. We examine a gate that changes the timing of calls to an existing within-paper evidence-search tool, measuring both overall inspection effort and the number of questions on which that effort reaches the target.

\section{Decision-Checkpoint Auditing}\label{sec:method}

\paragraph{Trajectory representation.} The agent searches for papers, queries evidence within a selected paper, or produces a final answer. For each turn $t$ with a model response, we define a checkpoint $c_t=(s_t,a_t,o_t,\mathrm{cost}_t)$. The state $s_t$ summarizes the maintained candidate list, prior queries, paper IDs with inspection attempts and retained evidence excerpts. This analytical summary is distinct from the model's input context and is not supplied as an additional message. The action $a_t$ represents the proposed action or, when the gate intervenes, the substituted evidence-search call; execution status is recorded separately. The observation $o_t$ contains returned paper IDs and evidence excerpts, while $\mathrm{cost}_t$ contains token use and latency. Search results enter the model context as tool messages, and older full-text returns are generally condensed into inspection summaries. Inference messages and checkpoints contain no ground-truth labels; benchmark metadata is associated with the trajectories only during subsequent evaluation.

\paragraph{Retrieval, exposure and inspection.} For a target paper $y^*$, \emph{retrieval} occurs when the backend includes $y^*$ among the top-$N$ results for a query. \emph{Exposure} occurs when the target identifier appears in a search-result list presented to the model, with its rank defined within that list. \emph{Inspection} occurs when an executed within-paper evidence-search call targets $y^*$. Exposure is measured from search results rather than subsequent repetitions of the identifier. Inspection includes unsuccessful calls and therefore denotes an attempt to obtain evidence, without establishing that adequate evidence was acquired.

\paragraph{Outcome categories.} After inference, target identities are matched to each trajectory and answer correctness is determined by the evaluator. Accepted answers are classified as \emph{Correct}, regardless of prior search or inspection. Incorrect answers with observable trajectories are classified as \emph{Incorrect, target inspected}; \emph{Incorrect, target exposed but not inspected}; or \emph{Incorrect, target neither exposed nor inspected}, with inspection taking precedence over exposure. API exceptions without trajectories form a separate unobservable category. This mutually exclusive classification describes outcomes without assuming a mandatory sequence of stages; exposure and inspection are also measured independently.

\paragraph{Exposure rank and subsequent decisions.} We characterize target visibility by its rank at first exposure and the number of search-result lists in which it appears. For incorrect answers with exposure but no inspection, we also measure whether at least one valid search, evidence-search or answer action follows first exposure. This continuation indicator establishes that another decision occurred after the target became visible, without estimating the remaining inspection budget or the counterfactual benefit of inspection.

\paragraph{Observation scope.} Exposure is determined from search observations and verified against retained tool messages; subsequent valid actions exclude parsing failures. Exposure records the target's appearance at any point in the trajectory, regardless of its later retention in the model context. Because final-message records omit some earlier evidence-search returns and checkpoints preserve excerpts rather than complete responses, the available trajectories support event-level analysis but not full reconstruction or semantic evaluation of every model input.

\section{Experimental Setting}\label{sec:setting}

\paragraph{Task and sample.} AutoResearchBench Deep requires identification of a single target arXiv paper or a ``no answer'' response. We evaluate five search conditions on all 540 answerable questions in its 600-question Deep split, excluding the 60 explicitly unsatisfiable questions. The resulting sample supports paper-identification analysis but does not assess abstention when no answer exists. A subset of 77 questions informed exploratory analysis and intervention design, so results on that subset are not independent confirmation. The study is restricted to Deep and a single model request alias. An episode denotes one execution of one condition on one question, and a trajectory is its sequence of actions and observations.

\paragraph{Retrieval environment.} The backend combines BM25~\citep{robertson2009bm25} retrieval over titles and abstracts from an indexed arXiv metadata snapshot~\citep{arxivsnapshot} containing 3.16 million records with BM25 retrieval over 7,703 cached full-text papers. Metadata retrieval uses $k_1{=}1.2$ and $b{=}0.75$, and the two rankings are combined by reciprocal-rank fusion~\citep{cormack2009rrf} with $c{=}60$. The full-text collection contains 3,853 targets from the broader benchmark corpus and 3,850 non-target papers matched by primary category and year; we denote this configuration \emph{+both}. Each full-text index entry contains at most 400,000 extracted characters. In contrast, within-paper evidence search uses the complete extracted text and returns up to three query-matched windows, each covering at most eight lines on either side of a match and truncated to 1,800 characters. Although ground-truth labels are withheld from model inputs, target identities are used to construct the full-text collection. Its target-enriched coverage can therefore influence both fused retrieval rankings and subsequent evidence access.

\paragraph{Search policies.} We use the AutoResearchBench inference framework with adapted tool prompts. The primary comparisons are keyword versus raw search and read-first versus keyword search. In \textsc{raw}, the agent is instructed to use the complete question verbatim for every search; this restriction is prompt-based rather than enforced by the execution framework. In \textsc{kw}, the agent formulates short, distinctive keyword queries and may reformulate them throughout the trajectory. Read-first retains the keyword rule and adds an inspection prompt and gate (Section~\ref{sec:actionable}). Two secondary conditions examine alternative query strategies: \textsc{decomp} divides the question into focused sub-questions, searches once for each and combines the evidence to select candidates; \textsc{adaptive} iteratively assesses candidates and reformulates queries using informative terms or titles until the evidence and candidate set stabilize. The suffix \texttt{-ft} denotes full-text-enabled policies on the shared +both backend. Appendix~\ref{app:implementation} specifies the query rules.

\paragraph{Execution and resource measurement.} All conditions use the request alias \texttt{gpt-5.6-luna}, the same retrieval backend and tool interfaces, and common nominal execution limits. The agent selects retrieval depth and evidence-search queries, subject to a candidate cap of 200 and a maximum of 30 rounds. The execution framework applies a soft context threshold of 110,000 estimated input tokens. Episodes reaching a round or context limit remain in the analysis. We measure Deep Accuracy@1 (pass@1), alongside search calls, evidence-search calls, token use and termination conditions. The category \emph{none} includes all final answers without an arXiv identifier, whether explicitly abstaining or terminating without an identified paper.

\paragraph{Answer evaluation.} The evaluator assesses whether the title of the first candidate in the final answer identifies the target paper; an absent candidate or explicit no-answer response scores zero on this answerable sample. Normalized paper IDs are used only in a separate scoring sensitivity analysis (Appendix~\ref{app:implementation}). The original runs do not fully specify the underlying model snapshots, effective reasoning settings or judge configurations, and the request alias alone does not establish a common model identity. Policy comparisons are consequently descriptive rather than verified same-model causal estimates. Re-evaluating all 2,700 answers with a uniform, documented title-based judge configuration reproduces every original correctness label, establishing label stability under that configuration while leaving judge validity and original model identity unresolved.

\section{Target Exposure and Failure Patterns}\label{sec:scale}

\begin{figure}[t]\centering
\includegraphics[width=\linewidth]{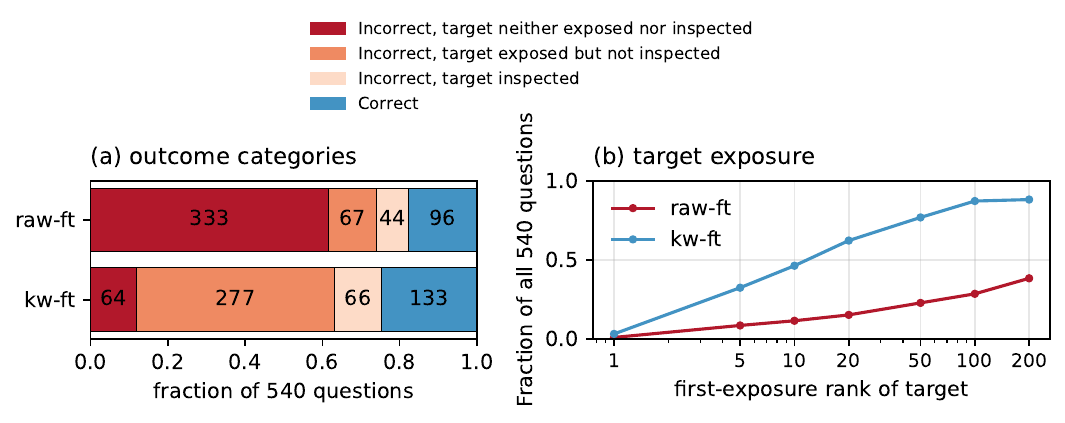}
\caption{Decision-checkpoint audit on all 540 answerable Deep questions (+both backend, request alias gpt-5.6-luna). (a) Correct answers take precedence; remaining questions are classified by inspection and exposure. Raw and keyword conditions differ in \emph{neither exposed nor inspected} (333$\to$64) and \emph{exposed, not inspected} (67$\to$277) failures, with Acc@1 of 17.8\% and 24.6\%. (b) Cumulative fraction of questions whose target first appears within a given rank of an observation; under \textsc{kw-ft} the target is exposed within rank 20 on 336/540 questions.}\label{fig:funnel}
\end{figure}

The keyword condition has fewer never-exposed failures and more exposed-but-uninspected failures than raw search (Figure~\ref{fig:funnel}, Table~\ref{tab:stage540}). Both conditions cover the same 540 answerable Deep questions; all outcome proportions use that 540-question denominator.

\begin{table}[t]\centering\small
\caption{Outcome categories across five conditions on the same 540 questions. Categories are mutually exclusive; all except Correct and Unobservable are incorrect outcomes. Exposure and inspection refer to the target paper. Exposure is also counted separately.}\label{tab:stage540}
\begin{tabular}{lrrrrr}\toprule
\multicolumn{6}{l}{(a) Outcome categories: counts, with accuracy in parentheses}\\
Condition & Correct $n$ (\%) & \shortstack{Neither exposed\\nor inspected} & \shortstack{Exposed, not\\inspected} & Inspected & Unobs.\\ \midrule
raw-ft & 96 (17.78) & 333 & 67 & 44 & 0 \\
kw-ft & 133 (24.63) & 64 & 277 & 66 & 0 \\
read-first & 133 (24.63) & 62 & 273 & 71 & 1 \\
decomp-ft & 138 (25.56) & 67 & 277 & 58 & 0 \\
adaptive-ft & 122 (22.59) & 68 & 282 & 68 & 0 \\ \bottomrule
\end{tabular}
\par\medskip
\begin{tabular}{lrrr}\toprule
\multicolumn{4}{l}{(b) Observed exposure and conditional uninspected failures}\\
Condition & Exposed $E/540$ (\%) & Uninspected failures & Later valid action in $U$\\
 & & $U/E$ (\%) & \\ \midrule
raw-ft & 207/540 (38.3) & 67/207 (32.4) & 67/67\\
kw-ft & 476/540 (88.1) & 277/476 (58.2) & 277/277\\
read-first & 477/540 (88.3) & 273/477 (57.2) & 273/273\\
decomp-ft & 473/540 (87.6) & 277/473 (58.6) & 277/277\\
adaptive-ft & 472/540 (87.4) & 282/472 (59.7) & 282/282\\ \bottomrule
\end{tabular}
\par\smallskip\raggedright
Accuracy is calculated over all 540 questions. Unobs. denotes the read-first episode without an observable trajectory because of an API exception; no exposure or inspection status is imputed for this episode. $U$ denotes exposed-but-uninspected failures, and $U/E$ is their proportion within each condition's exposed subgroup. These subgroups differ across conditions. Exposure is measured independently of the mutually exclusive outcome categories. A subsequent valid action indicates continuation after exposure, without establishing the remaining inspection budget. Appendix~\ref{app:aux} provides supplementary raw/keyword comparisons.
\end{table}

\begin{table}[t]\centering\small
\caption{Per-question outcome transitions raw-ft $\to$ kw-ft on the same 540 questions. All categories except Correct are incorrect outcomes; exposure and inspection refer to the target paper.}\label{tab:trans540}
\begin{tabular}{p{0.25\linewidth}rrrr}\toprule
raw $\downarrow$ \quad kw $\rightarrow$ & \shortstack{Neither exposed\\nor inspected} & \shortstack{Exposed, not\\inspected} & Inspected & Correct \\ \midrule
Neither exposed nor inspected (333) & 52 & \textbf{182} & 35 & 64 \\
Exposed, not inspected (67) & 4 & 46 & 8 & 9 \\
Inspected (44) & 6 & 13 & 12 & 13 \\
Correct (96) & 2 & 36 & 11 & 47 \\ \bottomrule
\end{tabular}
\end{table}

\begin{samepage}
\paragraph{Paired outcome transitions.} Keyword search has fewer failures without target exposure (62\%$\to$12\%) and an accuracy advantage of 6.85 percentage points over raw search. The paired comparison reveals that 182 of the 333 questions with neither exposure nor inspection under \textsc{raw} instead fail after exposure without inspection under \textsc{kw} (Table~\ref{tab:trans540}). Conversely, 36 questions answered correctly under \textsc{raw} become exposed-but-uninspected failures under \textsc{kw}.
\par\end{samepage}

\begin{samepage}
Target exposure frequently precedes further search decisions. All 277 keyword failures after exposure without inspection contain a subsequent valid action. Among the 476 keyword episodes with target exposure, 335 first expose the target by turn 3, and the median number of subsequent search, evidence-search or answer actions is 13. These observations establish continuation after exposure but do not quantify unused budget. Execution limits are also common in this failure category: 167 of 277 keyword episodes terminate at a token or turn limit, compared with 20 of 67 under \textsc{raw}.
\par\end{samepage}

\begin{samepage}
\paragraph{First-exposure rank and subsequent inspection.} Among the 336 keyword questions whose target first appears within the top 20, 184 (54.8\%) have no subsequent target inspection attempt. Even at ranks 1--5, 83 of 175 questions (47.4\%) have no subsequent attempt (Table~\ref{tab:rank}). These groups include both correct and incorrect answers, showing that uninspected targets also occur near the top of the result list. Appendix~\ref{app:aux} derives a separate lower bound restricted to incorrect answers.

\par\end{samepage}
\begin{table}[t]\centering\small
\caption{Keyword first-exposure rank, subsequent target inspection attempts and final correctness. All counts are numbers of questions. Percentages use the exposed questions in each rank group as denominators. ``Later target inspection'' requires an executed target call on a strictly later turn.}\label{tab:rank}
\setlength{\tabcolsep}{4pt}
\begin{tabular}{lrrrr}\toprule
First-exposure rank & 1--5 & 6--20 & 21--50 & 51--200\\ \midrule
Exposed questions $n$ & 175 & 161 & 79 & 61\\
Later target inspection, $n$ (\%) & 92 (52.6) & 60 (37.3) & 26 (32.9) & 13 (21.3)\\
Correct answers, $n$ (\%) & 65 (37.1) & 39 (24.2) & 21 (26.6) & 8 (13.1)\\ \bottomrule
\end{tabular}
\par\smallskip\raggedright
The 476 questions with target exposure are grouped by rank at first exposure. All 191 questions with a target inspection attempt have their attempt on a later turn. Percentages use the number of exposed questions in each rank group, and correctness follows the primary title-based evaluation. Outcomes are measured through episode termination. Because rank, exposure timing and subsequent trajectory length are uncontrolled, the associations do not identify a causal effect of rank.
\end{table}

\begin{samepage}
\section{Read-First Gate: Inspection and Answer Outcomes}\label{sec:actionable}

The prevalence of failures after exposure without inspection motivates examination of a read-first policy. This policy inserts a within-paper evidence-search attempt before another search or final answer whenever an eligible unread candidate remains. We compare its target inspection coverage, answer correctness and resource use with keyword search, distinguishing additional calls across all candidates from attempts directed at the target.

\par\end{samepage}
\begin{table}[t]\centering\small
\caption{Paired correctness and questions with a target inspection attempt. Gains and losses refer to the left-to-right policy change.}\label{tab:paired}
\begin{tabular}{lrrrrr}\toprule
\multicolumn{6}{l}{(a) Correctness on all $N=540$ questions}\\
Comparison & Both correct & Gain & Loss & Both wrong & $\Delta$Acc. (pp)\\ \midrule
raw $\to$ kw & 47 & 86 & 49 & 358 & +6.85\\
kw $\to$ read-first & 73 & 60 & 60 & 347 & 0.00\\
kw $\to$ decomp & 76 & 62 & 57 & 345 & +0.93\\
kw $\to$ adaptive & 69 & 53 & 64 & 354 & $-2.04$\\ \midrule
\multicolumn{6}{l}{(b) Target inspection attempt: kw $\to$ read-first, $N=539$ jointly observable}\\
Event & Both & Only read-first & Only kw & Neither & Unobs. pair\\ \midrule
Target inspection attempt & 133 & 66 & 57 & 283 & 1$^*$\\ \bottomrule
\end{tabular}
\par\smallskip\raggedright
$^*$The read-first API exception is included in the 540 answer outcomes but excluded from the 539 observable inspection pairs; its keyword counterpart contains a target inspection attempt. Target inspection is observed on 191 keyword and 199 read-first questions overall, or 190 and 199 within the paired observable sample. The latter difference is a net increase of nine across 123 changes in inspection status, whereas correctness changes on 120 questions. Every episode with a target inspection attempt contains at least one nonempty target excerpt returned on the same turn; this episode-level measure does not imply success for every call.
\end{table}

\begin{samepage}
\paragraph{Read-first policy.} After each paper search, the gate requires an inspection attempt before a subsequent search or final answer if an eligible candidate remains. A candidate is eligible when it has a valid normalized arXiv identifier and has not previously been the subject of an inspection attempt; unsuccessful attempts also remove eligibility. If the agent proposes another search or a final answer, the gate substitutes an evidence-search call on the highest-ranked eligible candidate in the latest result list, using the original question. This call consumes the current turn, after which the agent selects a new action. Any agent-initiated evidence-search call satisfies the gate, regardless of candidate rank or query. With no eligible candidate, the proposed action proceeds. Thus, although the prompt requests inspection of the highest-ranked unread candidate, the executed policy permits other agent-initiated inspections. The gate uses no ground-truth labels and remains subject to context and turn limits; limit-terminated episodes are included in all outcome and resource analyses.
\par\end{samepage}

Keyword and read-first each answer 133 of 540 questions correctly. They answer 73 questions correctly in both conditions; 60 are correct only under keyword search and 60 only under read-first (Table~\ref{tab:paired}). Read-first has more evidence-search calls per question than keyword search (9.63 versus 7.56), but recorded target inspection attempts occur on 199 questions under read-first and 191 under keyword search (Table~\ref{tab:resources}). Among the 539 questions with observable trajectories in both conditions, the corresponding numbers of questions with a target inspection attempt are 199 and 190.

\begin{samepage}
\paragraph{Sources of target inspection.} The 539 observable read-first trajectories contain 5,198 evidence-search calls: 1,720 forced by the gate and 3,478 initiated by the agent (Table~\ref{tab:resources}). Gate-forced calls reach the target on 22 questions and agent-initiated calls on 187, with an overlap of ten. Twelve questions therefore have target inspection only through forced calls, while 177 have it only through agent-initiated calls. These within-policy counts describe how inspection occurs; the paired comparison in Table~\ref{tab:paired} measures differences between policies. Every observed episode with target inspection contains at least one nonempty target excerpt associated with an executed call, although individual calls can fail. Appendix~\ref{app:implementation} defines this excerpt measure.
\par\end{samepage}

\paragraph{Uncertainty and sensitivity to exploratory questions.} The paired accuracy difference between keyword and raw search is +6.85 percentage points, with a 95\% question-bootstrap interval of [2.59, 11.11]. The read-first--keyword difference is 0.00 points, with an interval of [-4.07, 4.07]. Resampling the 529 target-identifier clusters gives corresponding intervals of [2.75, 11.05] and [-4.03, 4.05]. Excluding the 77 exploratory questions leaves 463 questions: keyword search has 74 gains and 43 losses relative to raw (+6.70 points; interval [2.16, 11.23]), while read-first has 52 gains and 50 losses relative to keyword (+0.43 points; [-3.89, 4.75]). These intervals describe variation across sampled questions or targets; they do not capture model variability, judge error or stochastic variation across repeated executions. Excluding exploratory questions is a sensitivity analysis rather than independent confirmation.

\paragraph{Inference resource use.} Following cost-aware agent evaluation~\citep{kapoor2024agentsmatter}, we compare resource use alongside answer accuracy. Read-first makes 27.4\% more recorded evidence-search calls and uses 4.8\% more inference tokens than keyword search, with identical aggregate accuracy (Table~\ref{tab:resources}).

\begin{table}[t]\centering\small
\caption{Recorded inference use and termination. Means use all 540 primary episodes per condition; termination columns are counts. Evidence-search calls include all candidate papers. The second panel separates call origins in the 539 observable read-first trajectories.}\label{tab:resources}
\setlength{\tabcolsep}{4pt}
\begin{tabular}{lrrrrrrr}\toprule
Condition & Searches & Evidence-search & Tokens & Normal & Context & Turn & API\\
 & /q & calls/q & /q (k) & & limit & limit & exception\\ \midrule
raw-ft & 3.61 & 8.43 & 844 & 386 & 131 & 23 & 0\\
kw-ft & 7.49 & 7.56 & 807 & 306 & 198 & 36 & 0\\
read-first & 7.51 & 9.63 & 846 & 301 & 172 & 66 & 1\\
decomp-ft & 7.38 & 7.43 & 762 & 290 & 205 & 45 & 0\\
adaptive-ft & 7.36 & 7.85 & 791 & 287 & 203 & 50 & 0\\ \bottomrule
\end{tabular}
\par\medskip
\begin{tabular}{lrrr}\toprule
Read-first call origin & \shortstack{Evidence-search\\calls} & \shortstack{Target inspection\\attempts} & \shortstack{Questions with a target\\inspection attempt}\\ \midrule
Harness-forced & 1,720 & 22 & 22\\
Agent-initiated & 3,478 & 381 & 187\\ \bottomrule
\end{tabular}
\par\smallskip\raggedright
Tokens denote cumulative recorded input and output ($\mathrm{k}=1000$); they measure neither peak context size nor billed cost. Context-limit termination includes the execution framework's soft threshold. Means use all 540 primary episodes, including the API exception, and reflect available usage records. They exclude the separate retry, evaluator calls and unrecorded infrastructure costs; missing usage is not interpreted as zero total cost.
\end{table}

\begin{samepage}
\paragraph{Sensitivity to the missing trajectory.} A separate stochastic retry of the read-first API exception exposes and inspects the target but still answers incorrectly. Substituting this retry leaves accuracy unchanged and assigns the previously unobservable episode to the inspected-but-incorrect category. The primary analysis retains the original episode, and the retry's resource use is excluded from primary-run costs.

\par\end{samepage}
\paragraph{Sensitivity to the scoring criterion.} Identifier-based matching identifies one target-title version mismatch classified as incorrect by the original title-based evaluator under keyword, read-first, decomposition and adaptive search, but not raw. Scoring solely by normalized target identity adds one correct answer in each of these four conditions, leaving the keyword--read-first accuracy difference at zero. Uniform title-based re-evaluation retains the original classifications. We therefore use the original title-based criterion for the primary results and report identifier agreement as a sensitivity analysis.

\section{Discussion and Limitations}
The results distinguish target visibility from subsequent inspection in scientific search. Keyword search has substantially fewer failures without exposure than raw search, yet many visible targets remain uninspected, including targets near the top of the result list. This pattern motivates examination of candidate selection and post-exposure resource allocation alongside retrieval. Inspection attempts followed by incorrect answers constitute a further group, but attributing those failures to reasoning requires evidence that the returned excerpts adequately address the question.

The read-first comparison shows why overall tool use and target-specific coverage should be measured separately. More evidence-search calls coincide with only a modest difference in the number of questions with target inspection, while paired comparisons reveal substantial changes in both inspection status and answer correctness. Identical aggregate accuracy consequently masks different question-level outcomes. Nonempty target excerpts establish that some content was returned during the inspection attempt, but do not establish its semantic sufficiency.

Several limitations constrain interpretation. The study covers one answerable benchmark subset in a target-enriched retrieval environment, and 77 questions informed intervention design. The original underlying-model identities and effective reasoning settings are incompletely specified. Moreover, without repeated executions within each condition, policy differences cannot be separated from stochastic trajectory variation; resampling questions or targets does not supply such replication. Finally, retained excerpts permit event-level measurement but not complete reconstruction or semantic assessment of every model input. The findings are therefore descriptive, with limited generalizability and no identified causal mechanism.

\begin{samepage}
\section{Conclusion}
Decision-checkpoint auditing provides an event-based account of target exposure, inspection attempts and answer correctness in scientific search. Across 540 answerable AutoResearchBench Deep questions in a target-enriched environment, keyword search has fewer failures without exposure or inspection than raw search, but more failures after exposure without inspection. Read-first makes 27.4\% more evidence-search calls than keyword search, while target inspection is observed on 199 rather than 191 questions. Both policies answer 133 questions correctly, with 60 gains and 60 losses in the paired comparison. Measuring these events alongside final accuracy reveals differences in search behavior and question-level outcomes that aggregate scores alone obscure.
\par\end{samepage}

\subsubsection*{AI use statement}
AI systems assisted with methodological and experimental design, code and implementation, qualitative analysis and interpretation of trajectories and results, figure preparation, and manuscript drafting and revision, under the direction of the authors. The authors are responsible for the final content of the manuscript.

\bibliography{refs}

@article{robertson2009bm25,
  author = {Robertson, Stephen and Zaragoza, Hugo},
  title = {The Probabilistic Relevance Framework: {BM25} and Beyond},
  journal = {Foundations and Trends in Information Retrieval},
  year = {2009},
  doi = {10.1561/1500000019}
}

@inproceedings{cormack2009rrf,
  author = {Cormack, Gordon V. and Clarke, Charles L. A. and B{\"u}ttcher, Stefan},
  title = {Reciprocal Rank Fusion outperforms Condorcet and individual Rank Learning Methods},
  booktitle = {Proceedings of the 32nd International ACM SIGIR Conference on Research and Development in Information Retrieval},
  pages = {758--759}, year = {2009}, doi = {10.1145/1571941.1572114}
}

@article{kapoor2024agentsmatter,
  author = {Kapoor, Sayash and Stroebl, Benedikt and Siegel, Zachary S. and Nadgir, Nitya and Narayanan, Arvind},
  title = {{AI} Agents That Matter},
  journal = {arXiv preprint arXiv:2407.01502}, year = {2024},
  url = {https://arxiv.org/abs/2407.01502}
}

@article{autoresearchbench,
  title={{AutoResearchBench}: Benchmarking {AI} Agents on Complex Scientific Literature Discovery},
  author={Lei Xiong and Kun Luo and Ziyi Xia and Wenbo Zhang and Jin-Ge Yao and Zheng Liu and Jingying Shao and Jianlyu Chen and Hongjin Qian and Xi Yang and Qian Yu and Hao Li and Chen Yue and Xiaan Du and Yuyang Wang and Yesheng Liu and Haiyu Xu and Zhicheng Dou},
  journal={arXiv preprint arXiv:2604.25256},
  year={2026},
  note={\url{https://github.com/CherYou/AutoResearchBench}}
}

@article{budgetcontrol,
  title={Inference-Time Budget Control for {LLM} Search Agents},
  author={Zhengru Fang and Senkang Forest Hu and Zhonghao Chang and Yu Guo and Yihang Tao and Hongyao Liu and Mengzhe Ruan and Jun Huang and Yuguang Fang},
  journal={arXiv preprint arXiv:2605.05701},
  year={2026}
}

@article{valuetree,
  title={Spend Less, Reason Better: Budget-Aware Value Tree Search for {LLM} Agents},
  author={Yushu Li and Wenlong Deng and Jiajin Li and Xiaoxiao Li},
  journal={arXiv preprint arXiv:2603.12634},
  year={2026}
}

@misc{arxivsnapshot,
  title={{arXiv} Dataset},
  author={{Cornell University}},
  howpublished={Kaggle, \url{https://www.kaggle.com/datasets/Cornell-University/arxiv}},
  note={CC0 metadata; local index documented September 2026; source version and download date unrecorded},
  year={n.d.}
}

@article{he2025pasa,
  title={{PaSa}: An {LLM} Agent for Comprehensive Academic Paper Search},
  author={He, Yichen and Huang, Guanhua and Feng, Peiyuan and Lin, Yuan and
          Zhang, Yuchen and Li, Hang and E, Weinan},
  journal={arXiv preprint arXiv:2501.10120},
  year={2025},
  url={https://arxiv.org/abs/2501.10120v2}
}

@inproceedings{ajith2024litsearch,
  title={{LitSearch}: A Retrieval Benchmark for Scientific Literature Search},
  author={Ajith, Anirudh and Xia, Mengzhou and Chevalier, Alexis and
          Goyal, Tanya and Chen, Danqi and Gao, Tianyu},
  booktitle={Proceedings of the 2024 Conference on Empirical Methods in Natural Language Processing},
  pages={15068--15083},
  doi={10.18653/v1/2024.emnlp-main.840},
  year={2024},
  url={https://aclanthology.org/2024.emnlp-main.840/}
}

@article{asai2024openscholar,
  title={{OpenScholar}: Synthesizing Scientific Literature with Retrieval-augmented {LMs}},
  author={Asai, Akari and He, Jacqueline and Shao, Rulin and Shi, Weijia and
          Singh, Amanpreet and Chang, Joseph Chee and Lo, Kyle and
          Soldaini, Luca and Feldman, Sergey and D'arcy, Mike and
          Wadden, David and Latzke, Matt and Tian, Minyang and Ji, Pan and
          Liu, Shengyan and Tong, Hao and Wu, Bohao and Xiong, Yanyu and
          Zettlemoyer, Luke and Neubig, Graham and Weld, Dan and Downey, Doug and
          Yih, Wen-tau and Koh, Pang Wei and Hajishirzi, Hannaneh},
  journal={arXiv preprint arXiv:2411.14199},
  year={2024},
  url={https://arxiv.org/abs/2411.14199}
}

@article{chen2025browsecompplus,
  title={{BrowseComp-Plus}: A More Fair and Transparent Evaluation Benchmark of Deep-Research Agent},
  author={Zijian Chen and Xueguang Ma and Shengyao Zhuang and Ping Nie and Kai Zou and Andrew Liu and Joshua Green and Kshama Patel and Ruoxi Meng and Mingyi Su and Sahel Sharifymoghaddam and Yanxi Li and Haoran Hong and Xinyu Shi and Xuye Liu and Nandan Thakur and Crystina Zhang and Luyu Gao and Wenhu Chen and Jimmy Lin},
  journal={arXiv preprint arXiv:2508.06600},
  year={2025}
}

@article{agenteval2026,
  title={{AgentEval}: {DAG}-Structured Step-Level Evaluation for Agentic Workflows with Error Propagation Tracking},
  author={Dongxin Guo and Jikun Wu and Siu Ming Yiu},
  journal={arXiv preprint arXiv:2604.23581},
  year={2026}
}

@misc{liang2026searchauditor,
  title={{SearchAuditor}: Auditing and Attributing Failures in Long-Horizon Search Agents},
  author={Zhixiang Liang and Yifei Liu and Yidan Huang and Haozhe Zhao and Beichen Huang and Jiaqi Wang and Nan Duan and Qiong Cao},
  year={2026},
  eprint={2608.05212},
  archivePrefix={arXiv},
  primaryClass={cs.AI},
  url={https://arxiv.org/abs/2608.05212}
}

@article{skarlinski2024paperqa,
  title={Language agents achieve superhuman synthesis of scientific knowledge},
  author={Skarlinski, Michael D. and Cox, Sam and Laurent, Jon M. and Braza, James D. and Hinks, Michaela and Hammerling, Michael J. and Ponnapati, Manvitha and Rodriques, Samuel G. and White, Andrew D.},
  journal={arXiv preprint arXiv:2409.13740},
  year={2024},
  url={https://arxiv.org/abs/2409.13740v1}
}

@article{ma2024agentboard,
  title={{AgentBoard}: An Analytical Evaluation Board of Multi-turn {LLM} Agents},
  author={Ma, Chang and Zhang, Junlei and Zhu, Zhihao and Yang, Cheng and Yang, Yujiu and Jin, Yaohui and Lan, Zhenzhong and Kong, Lingpeng and He, Junxian},
  journal={arXiv preprint arXiv:2401.13178},
  year={2024},
  url={https://arxiv.org/abs/2401.13178v1}
}

@inproceedings{lu2025toolsandbox,
  title={{ToolSandbox}: A Stateful, Conversational, Interactive Evaluation Benchmark for {LLM} Tool Use Capabilities},
  author={Lu, Jiarui and Holleis, Thomas and Zhang, Yizhe and Aumayer, Bernhard and Nan, Feng and Bai, Haoping and Ma, Shuang and Ma, Shen and Li, Mengyu and Yin, Guoli and Wang, Zirui and Pang, Ruoming},
  booktitle={Findings of the Association for Computational Linguistics: NAACL 2025},
  pages={1160--1183},
  year={2025},
  doi={10.18653/v1/2025.findings-naacl.65},
  url={https://aclanthology.org/2025.findings-naacl.65/}
}

@inproceedings{wadden2020scifact,
  title={Fact or Fiction: Verifying Scientific Claims},
  author={Wadden, David and Lin, Shanchuan and Lo, Kyle and Wang, Lucy Lu and van Zuylen, Madeleine and Cohan, Arman and Hajishirzi, Hannaneh},
  booktitle={Proceedings of the 2020 Conference on Empirical Methods in Natural Language Processing (EMNLP)},
  pages={7534--7550},
  year={2020},
  doi={10.18653/v1/2020.emnlp-main.609},
  url={https://aclanthology.org/2020.emnlp-main.609/}
}

@article{ru2024ragchecker,
  title={{RAGChecker}: A Fine-grained Framework for Diagnosing Retrieval-Augmented Generation},
  author={Ru, Dongyu and Qiu, Lin and Hu, Xiangkun and Zhang, Tianhang and Shi, Peng and Chang, Shuaichen and Jiayang, Cheng and Wang, Cunxiang and Sun, Shichao and Li, Huanyu and Zhang, Zizhao and Wang, Binjie and Jiang, Jiarong and He, Tong and Wang, Zhiguo and Liu, Pengfei and Zhang, Yue and Zhang, Zheng},
  journal={arXiv preprint arXiv:2408.08067},
  year={2024},
  url={https://arxiv.org/abs/2408.08067v2}
}
\bibliographystyle{plainnat}

\clearpage
\appendix
\section{Supplementary Raw and Keyword Comparisons}\label{app:aux}
Table~\ref{tab:aux} summarizes additional exposure and termination statistics for raw and keyword search. Of the 476 keyword questions with target exposure, 336 first expose the target within rank 20 and 277 are answered incorrectly without target inspection. Their intersection therefore contains at least $336+277-476=137$ questions. This lower bound concerns incorrect answers; the 184 uninspected top-20 episodes in the main text include both correct and incorrect answers.
\begin{table}[h]\centering\small
\caption{Supplementary exposure, outcome and resource statistics for raw and keyword search on 540 questions per condition.}\label{tab:aux}
\begin{tabular}{lrr}\toprule
Measure & raw-ft & kw-ft\\ \midrule
Never-exposed failures (\%) & 61.7 & 11.9\\
Exposed, uninspected failures (\%) & 12.4 & 51.3\\
Inspected, incorrect failures (\%) & 8.1 & 12.2\\
Correct (\%) & 17.8 & 24.6\\
Search-result lists containing the target per question & 1.0 & 3.8\\
Uninspected-failure stops: normal / turn / context limit & 47 / 4 / 16 & 110 / 24 / 143\\
Searches / evidence-search calls per question (rounded) & 3.6 / 8.4 & 7.5 / 7.6\\
\shortstack[l]{Episodes ending without an arXiv ID\\in the final answer} & 238 & 73\\ \bottomrule
\end{tabular}
\end{table}

\section{Measurement and Scoring Procedures}\label{app:implementation}

\paragraph{Identifier normalization and event definitions.} Paper identifiers are normalized by removing whitespace, repository prefixes and version suffixes, accepting both modern and legacy arXiv formats. For modern identifiers dated January 2015 or later, four-digit sequences are padded to five digits. Normalization standardizes identifier comparisons without independently validating paper identity. First exposure is the earliest appearance of the target in a search-result list, with rank counted from one. Inspection requires an executed call directed at the target; proposed but unexecuted actions do not qualify. Repeated identifiers outside search observations do not constitute new exposure. Subsequent decisions include valid search, within-paper evidence-search and answer actions after exposure through episode termination.

\paragraph{Returned target excerpts.} Each checkpoint retains up to the first 1,800 characters of the latest evidence-search response. We identify a nonempty target excerpt when the response to an executed target call identifies that paper and contains query-matched text with a source line number. Empty responses, errors and returns without a match are excluded. Episode-level coverage requires at least one such excerpt, so it does not imply success for every call or sufficient evidence for the question. Because checkpoints retain only part of each response, this measure concerns recorded excerpts rather than complete tool returns or model inputs.

\paragraph{Primary outcome scoring.} The evaluator compares the metadata title of the first candidate in the final answer with the target title in a single pass. Absent or empty titles and no-answer responses score zero in the answerable sample. The judge assesses same-paper identity, allowing truncation and differences in punctuation, capitalization or subtitles when identity is preserved; it does not independently verify the question's constraints. The original evaluation extracts a Boolean title-match decision from the judge response, with a fallback based on the presence of an affirmative keyword when the response cannot be parsed. Normalized target-identifier agreement is evaluated separately and does not override the primary labels.

\paragraph{Sensitivity re-evaluation.} All 2,700 answers are re-evaluated using the original title prompts and a common judge configuration: the model request alias \texttt{gpt-5.6-luna}, temperature zero and medium reasoning effort. The service does not provide an immutable underlying-model snapshot or enforce the requested completion cap. The sensitivity evaluation accepts only explicitly parsed Boolean decisions and leaves unresolved responses unscored, without using original labels to fill missing decisions. All answers are resolved and agree with the primary labels. This agreement supports label stability under the specified configuration; the original judge configuration remains incompletely documented.

\paragraph{Query policies and gate eligibility.} The RAW prompt requires the complete question verbatim in every search. It allows changes in retrieval depth, evidence inspection and termination, while prohibiting query shortening, paraphrasing, term addition, title substitution and decomposition. KW instructs the agent to formulate short, distinctive keyword queries before each search, allowing later queries to differ. These are continuing prompt instructions; the execution framework requires nonempty queries but does not enforce verbatim adherence in RAW. Under read-first, eligibility is restricted to candidates in the latest result list that have a valid normalized identifier and no prior inspection attempt. If no candidate qualifies, the next search or answer action proceeds without substitution. Continuation after target exposure is measured by the presence of a subsequent valid search, inspection or answer action.

\paragraph{Aggregate inspection outcomes.} The keyword and read-first conditions have the following aggregate outcomes, complementing the breakdown by call origin in Table~\ref{tab:resources}.
\begin{center}\small
\begin{tabular}{lrr}\toprule
Outcome & Keyword & Read-first\\ \midrule
Accepted answers & 133/540 & 133/540\\
\shortstack[l]{Questions with a target inspection\\attempt (all 540)} & 191 & 199\\
\shortstack[l]{Questions with a target inspection\\attempt (539 observable pairs)} & 190 & 199\\
Unobservable trajectories & 0 & 1\\ \bottomrule
\end{tabular}
\end{center}

\paragraph{Inspection origin and unsuccessful returns.} Evidence-search calls are classified as gate-forced when an executed inspection replaces the agent's proposed search or answer action, and as agent-initiated when the agent itself requests inspection. Classification matches each executed action to its corresponding turn and paper identifier. Among the 539 observable read-first trajectories, 1,720 calls are gate-forced and 3,478 are agent-initiated. The former reach the target on 22 questions through 22 calls; the latter reach it on 187 questions through 381 calls. Ten questions belong to both groups, yielding twelve with forced-only attempts and 177 with agent-only attempts, for 199 questions overall. The 5,198 calls correspond to 9.63 per question over the full 540-question sample. Gate-forced calls include 15 errors and no explicit no-match returns, while agent-initiated calls include 15 errors and three no-match returns. The episode without a trajectory remains unobservable. These counts describe executed calls and recorded returns, without establishing a counterfactual increase in inspection or semantic evidence sufficiency.

\paragraph{Paired bootstrap estimation.} We quantify uncertainty in accuracy differences using 100,000 bootstrap samples for each policy comparison and resampling scheme. For question-level resampling, each question contributes a paired correctness difference of $-1$, $0$ or $1$, and bootstrap samples draw from the empirical distribution of these differences. For cluster-level resampling, questions are grouped by their exact target-identifier set. Each sample draws the original number of clusters uniformly with replacement and retains all questions within each selected cluster, including their multiplicity when a cluster is sampled more than once. The accuracy difference is the sum of paired differences divided by the sampled question count, so clusters have equal selection probability but contribute in proportion to their question counts. The 95\% percentile intervals use the 2.5th and 97.5th percentiles with linear interpolation and are expressed in percentage points. Both schemes are applied to the full sample and after excluding exploratory questions. Resampling uses observed answer outcomes and does not repeat agent executions.

\paragraph{Event-label consistency.} Automated comparisons with AI-assisted inspection of 40 stratified trajectories and five additional anomaly records found no discrepancies between the recorded events and their summaries. This check assesses agreement with the available trajectories rather than semantic evidence sufficiency.

\paragraph{Exploratory assessment of semantic support.} An AI-assisted constraint-level review examined 26 retained excerpts from ten selected records, rather than complete tool returns. The review identified partial support and one unresolved ambiguity concerning the direction of a comparison. These exploratory findings do not establish a benchmark error and do not provide independent human validation, demonstrate full evidence sufficiency, or attribute failures to reasoning.

\end{document}